\documentclass{article}

\usepackage[nohyperref,accepted]{icml2018}

\usepackage[utf8]{inputenc} 
\usepackage[T1]{fontenc}    
\usepackage{url}            

\usepackage{breakurl}
\usepackage[breaklinks]{hyperref}
\usepackage{booktabs}        

\usepackage{amsmath,amsfonts,amsthm,bm} 
\usepackage{algorithm}
\usepackage{nameref,hyperref}

\usepackage{graphicx}
\usepackage{subcaption}
\usepackage{microtype}      
\usepackage{float}

\title{Sparse Unsupervised Capsules Generalize Better}

\author{David Rawlinson\\
Incubator 491 \\
\texttt{dave@agi.io}
\and
Abdelrahman Ahmed\\
Incubator 491 \\
\texttt{abdel@agi.io}
\and
Gideon Kowadlo\\
Incubator 491\\
\texttt{gideon@agi.io}
}

\begin{document}

\maketitle

\begin{abstract}
We show that unsupervised training of latent capsule layers using only the reconstruction loss, without masking to select the correct output class, causes a loss of equivariances and other desirable capsule qualities. This implies that supervised capsules networks can’t be very deep. Unsupervised sparsening of latent capsule layer activity both restores these qualities and appears to generalize better than supervised masking, while potentially enabling deeper capsules networks. We train a sparse, unsupervised capsules network of similar geometry to \cite{Sabour2017} on MNIST \cite{LeCun1998} and then test classification accuracy on affNIST\footnote{\url{https://www.cs.toronto.edu/~tijmen/affNIST/}} using an SVM layer. Accuracy is improved from benchmark 79\% to 90\%.
\end{abstract}

\section{Introduction}

Capsules networks \cite{Sabour2017, Hinton2018} are built from blocks called Capsules that use a mechanism called ``dynamic routing'' to transiently build an ensemble of modules that respond to current input. It has been shown that training of residual networks \cite{Szegedy2017} also produces transient ensembles of modules to handle specific input \cite{Veit2016}, but the mechanism in capsules networks is more explicit and more powerful, because it considers feedback data.

Dynamic routing is a consensus mechanism between capsules throughout a multi-layer hierarchy. Since capsules must agree on the ``pose'' (or ``instantiation parameters'' \cite{Hinton2011}) of the entities being modelled, it is conceptually similar (but not numerically equivalent) to the concept of inference via message-passing-algorithms such as Belief Propagation \cite{Yedidia2003}, that can output marginal distributions over hidden variables at nodes in a graphical model. Routing can also be interpreted as means of integrating feedback, which has been shown to generally improve network performance in metrics such as robustness to occlusion and noise \cite{Shrivastava2016}. 

\subsection{Equivariances}
The output of a capsule is a vector, not a scalar. It represents a parameterization of its input in terms of some latent variables implicitly discovered and modelled by specialized capsules. An important consequence of this learned specialization is that a capsule’s output is equivariant, meaning that a range of transformations or variations of the input can be described by continuous changes in the output vector. 

The capsules concept was motivated by deficiencies in the representational capabilities of convolutional networks and traditional autoencoders \cite{Hinton2011}. However, the first paper to demonstrate automatic discovery of equivariances in latent variables was \cite{Sabour2017}. In this work, capsules must produce output that can be transformed into a consensus parameterization, or be ignored by exclusion from the parse-tree: predicting this transformation is the consensus mechanism and the parse tree is the resulting ensemble. 

\begin{figure}[H]  
  \centering
    \includegraphics[width=0.4\textwidth]{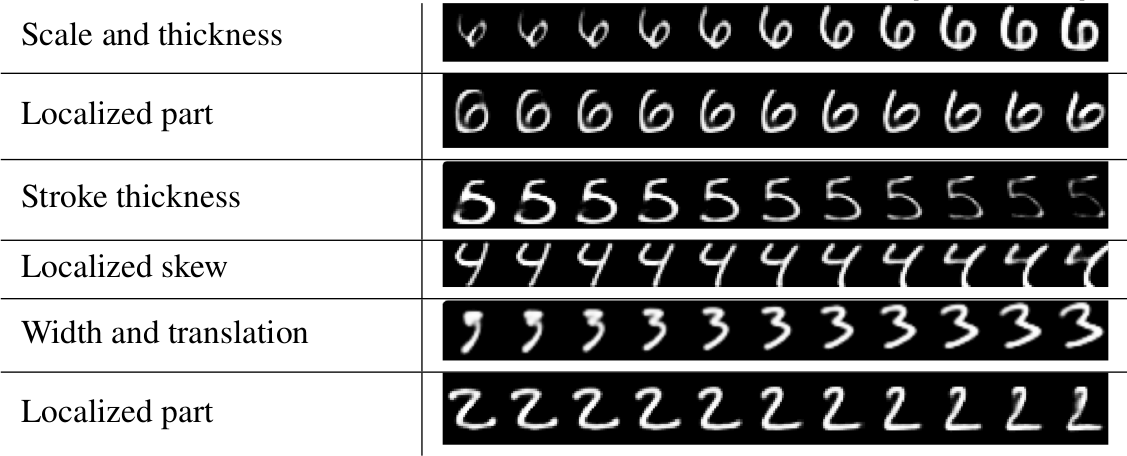}
  \caption{Sample equivariances modelled by the original capsules algorithm, reproduced from \cite{Sabour2017}.}\label{fig:1}
\end{figure}

One of the key qualities anticipated of capsules networks is that equivariance may prove to be a more generalizable representation than conventional network outputs, that usually strive for invariance. Early evidence of this is provided in \cite{Sabour2017} in which they trained a capsules network on the MNIST dataset \cite{LeCun1998} and then tested on the affNIST (affine-transformed MNIST) dataset, achieving 79\% classification accuracy. This surprising result is believed to be due to the capsules network being inherently able to describe affine-transformed digits via its equivariant representation. Further evidence of this is provided by demonstrations of the effect of varying individual parameters of capsules' output - see figure \ref{fig:1}: The network learned to capture properties such as stroke width and digit skew.

\subsection{Unsupervised Learning}
Sparse unsupervised learning was certainly considered during the early development of capsules networks, for example in Tieleman’s thesis \cite{Tieleman2014}, but is not a feature of recent works.

In \cite{Sabour2017, Hinton2018}, routing is unsupervised and does not directly train any weights. However, by modulating capsules' output, routing does affect the outcome of supervised discriminatory training. Two loss functions are used; one for digit classification, applied at the deepest capsule layer; and a reconstruction loss applied to a decoder network and backpropagated through the capsules layers. The reconstruction loss function is subject to a one-hot mask vector that trains a single winning capsule to represent the input.

In this paper, we create a fully unsupervised capsules network and investigate its qualities. Several factors motivate this. In particular, unsupervised learning does not require labelled training data, which is often costly and difficult to produce in quantity. 

It is generally believed that unsupervised training or pre-training produces more regularized models than supervised discriminatory training \cite{Erhan2010}. This is because unsupervised learning only attempts to fit the statistics of the observed data, rather than optimizing a separate function. This can produce models that generalize better, although not in all cases.

Discriminatory function optimization with backpropagation of supervised losses improves task-specific network performance by optimizing weights and decision boundaries in all layers simultaneously. The entire model is optimized for the discriminatory task. Negative consequences of this optimization include susceptibility to adversarial attacks \cite{Goodfellow2014,Kurakin2016}.

Despite a thorough search \cite{Bengio2016,Balduzzi2015}, researchers have failed to find a biological equivalent to credit assignment via error-gradient backpropagation. We would like to better understand and emulate the capabilities of natural neural networks \cite{Rawlinson2017}.

\section{Experiments}
We modified the architecture of \cite{Sabour2017} (hereafter, SUPCAPS) to use unsupervised learning. We also use the same notation and variables. Except where stated, our implementation is identical to SUPCAPS. We will call our modified algorithm SPARSECAPS. The more recent implementation of Capsules by \cite{Hinton2018} introduced an improved Expectation-Maximization routing algorithm (EMCAPS) that supersedes the dynamic routing used in this paper and Sabour et al. However, we believe the conclusions of this paper are not substantially affected by these changes as both prior works use unsupervised routing and supervised discriminatory training of network weights.

SUPCAPS uses two loss functions, a margin loss on classification accuracy and a reconstruction loss. The latter utilizes a decoder network on the output of the deepest capsules layer, resulting in an autoencoder architecture. 

The deepest capsule layer in SUPCAPS is termed the ‘DigitCaps’ layer, because there is one capsule per digit. Classification is performed using the output of the DigitCaps layer by finding the digit-capsule with the largest activity vector norm.

Less obviously, the DigitCaps output is masked during training with a one-hot vector such that only the ‘correct’ digit capsule has nonzero output. Masking assigns a meaning to each digit-capsule, forcing it to represent the various forms of a single digit. Masking also has the drawback that the interpretation of each digit-capsule must be known and determined by the algorithm designer in advance. This is not possible in an unsupervised model. Instead, we rename the deepest layer of capsules to be latent-capsules, representing some latent variables to be discovered in the data.

\subsection{Dense Unsupervised Capsules}
\begin{figure}  
  \centering
  
\begin{subfigure}[b]{0.4\textwidth}
   \includegraphics[width=1\linewidth]{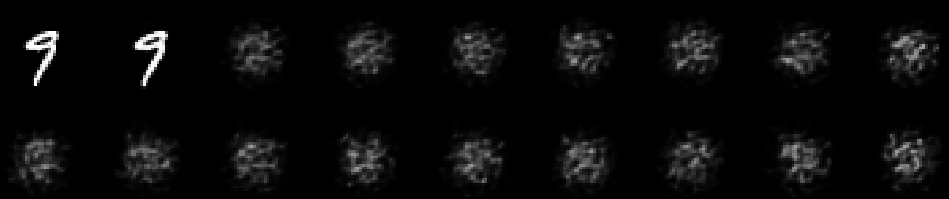}
   \caption{From top-left: Input image, reconstruction using all capsules, and then reconstruction using each capsule individually.}
\end{subfigure}

\vspace{1pc}

\begin{subfigure}[b]{0.4\textwidth}
   \includegraphics[width=1\linewidth]{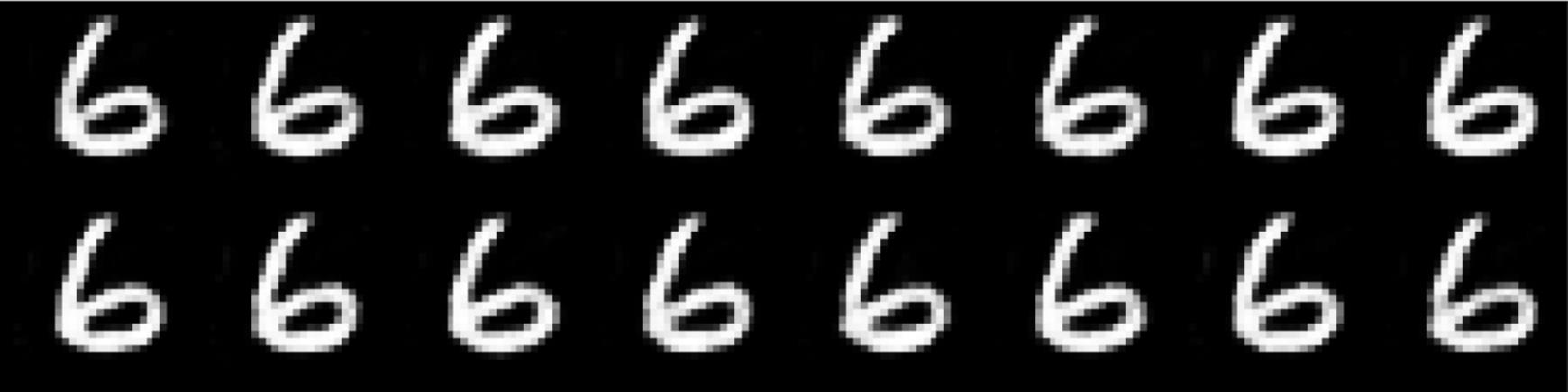}
   \caption{Reconstructions using a ``leave one out'' approach.}
\end{subfigure}

\vspace{1pc}

\begin{subfigure}[b]{0.4\textwidth}
   \includegraphics[width=1\linewidth]{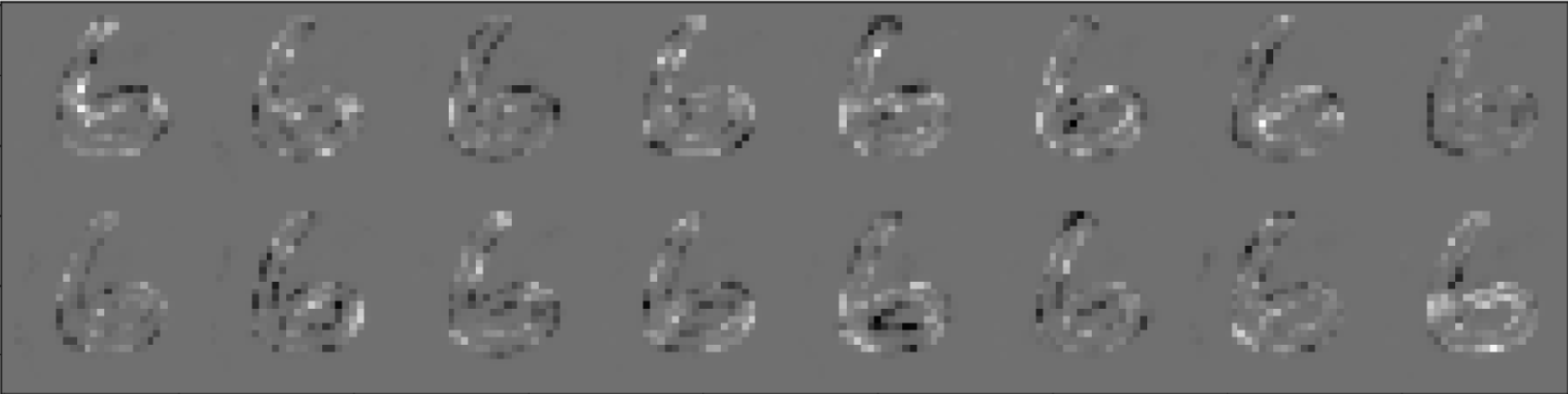}
   \caption{Reconstruction changes when each capsule is excluded.}
\end{subfigure}
  
  \caption{Digit reconstruction from unsupervised capsules without sparse training. Although the combined reconstruction is good, note the absence of any spatial correlation in individual capsules' output. All capsules contribute a little to the output, so excluding any one has little effect. No clique of capsules preferentially responds to any input. Effectively, the whole layer behaves as an autoencoder.}\label{fig:2}
\end{figure}

\begin{figure}  
  \centering
    \includegraphics[width=0.45\textwidth]{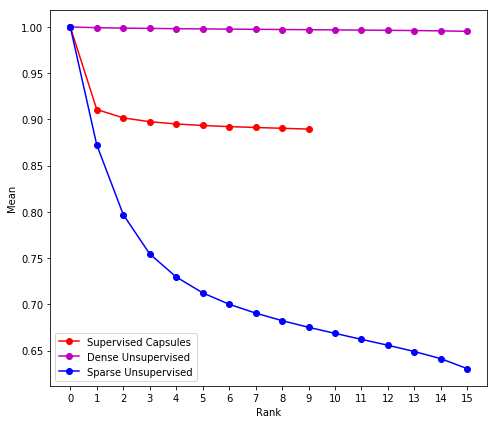}
  \caption{Ranked dynamic routing coefficients (averaged over 10,000 images). We rank latent capsules by their aggregate routing coefficients $\psi_{j}$ (see text). The ranked values are scaled by $\psi_{j}$ of the capsule with rank 0. Dynamic routing is intended to produce a parse-tree from the available capsules by assigning high routing coefficients to consensus capsules and lower coefficients to other capsules. Without sparse training, unsupervised capsules all have similar coefficients, suggesting that the routing mechanism is no longer able to build a discriminating parse-tree. Adding sparse training restores the ability to select capsules via routing.}\label{fig:4}
\end{figure}

Since SUPCAPS was always trained with an unsupervised reconstruction loss function, our first attempt to make the entire network unsupervised was simply to remove the mask and margin loss functions. Interestingly, this had the effect of turning the entire network into an autoencoder, with good reconstruction loss but a breakdown of capsule qualities. We observed (see Figure \ref{fig:2}) that under these conditions all capsules participate in all outputs (rather than selectively responding to different inputs), that capsules ceased to discover equivariances, that the routing consensus mechanism was no longer functional (Figure \ref{fig:4}) and that generalization from MNIST to affNIST data failed (see results). These consequences are very unfortunate, because they suggest that deep capsules networks will not work without supervised masking of latent capsules to force them to represent particular latent entities. Although this masking is only applied to the deepest capsule layer, the effects back-propagate to shallower capsule layers. It is likely that within a few layers, the degenerate capsule behaviour observed without masking would occur even in a supervised capsules network. 


\subsection{Sparse Unsupervised Capsules}
Intuitively, to restore the original capsule behaviour it is necessary to allow capsules to specialize; each latent-capsule should discover a subset of the input space that it can model and parameterize efficiently. This is another way of saying that latent-capsules within a layer should be sparsely active. 

The effects of sparseness have been studied thoroughly (see \cite{Olshausen1997} for a good introduction). More recently, and at scale, \cite{Le2012} explored sparse training of a hierarchical, unsupervised representation on internet videos. While the potential benefits of sparseness are relatively well understood, there is less agreement on the most efficient and effective training methods.

Several complex but principled methods of sparse coding have been described, including sparsity penalties \cite{Lee2007,Nair2009}, MOD \cite{Engan1999} and K-SVD \cite{Aharon2005}. We took inspiration from the simple k-sparse scheme of \cite{Makhzani2013}, in which $k$ winning cells with the highest hidden-layer activity are allowed to contribute to the output of an autoencoder, and the rest are masked. The only nonlinearity in their algorithm is sparse masking.

To sparsen the output of the latent capsules layer, we allowed the existing iterative dynamic routing to determine which latent capsules had the largest inbound routing coefficients $c_{ijk}$ where $c_{ij}$ is as defined in Sabour et al and $k$ is an index over the current batch. The primary capsules layer is convolutional, so the matrix of routing coefficients in the latent capsules layer has dimension $[ K, W, H, P ]$ where batch size $K=128$, the convolutional width and height of the primary capsules layer $W=6$ and $H=6$, and $P=32$ (the depth of the primary capsules layer, in capsules). Since routing produces a parse-tree ensemble of capsules, we are only interested in the largest routing coefficient at each convolutional location $x,y$ in the primary capsules layer. To this end we take the max over $P$ and sum over $W$ and $H$ giving $\psi_{jk}$, the total routing support for each latent capsule $j$ in each batch sample $k$. 

SPARSECAPS uses $L=16$ latent capsules to obtain a similar network size to SUPCAPS without pre-determining that the network should model 10 unique digits. We generated a sparse-mask value $m_{jk}$ for each latent capsule $j$ by ranking capsule support $\psi_{j}$ in descending order (i.e. max support yields rank 0) and then passing the relative rank into an exponential function (equation \ref{eq:4}). Mask values less than 0.01 are forced to zero for numerical stability.

The Hadamard (elementwise) product of the mask $m$ with latent capsules' output $v$ is the only sparsening step required. Using $L=16$ latent capsules and $\gamma = 12$, approximately 1 capsule is fully active and another two are partially active. 

\subsection{Online Lifetime Sparsity Constraint}
Analysis of this algorithm showed that only a few capsules were consistently active for all inputs, losing the specialization required for the discovery and parameterization of equivariances. To combat this we added a ``lifetime-sparsity constraint'' \cite{Makhzani2015} to ensure that all capsules occasionally participate in the output. For ease of implementation we used an ad-hoc, online boosting system to increase the output of rarely used capsules and reduce the output of overused capsules when rank-zero-frequency is outside a specified range. The boosting step is shown in equation \ref{eq:1}.

\begin{figure}  
  \centering
    \includegraphics[width=0.45\textwidth]{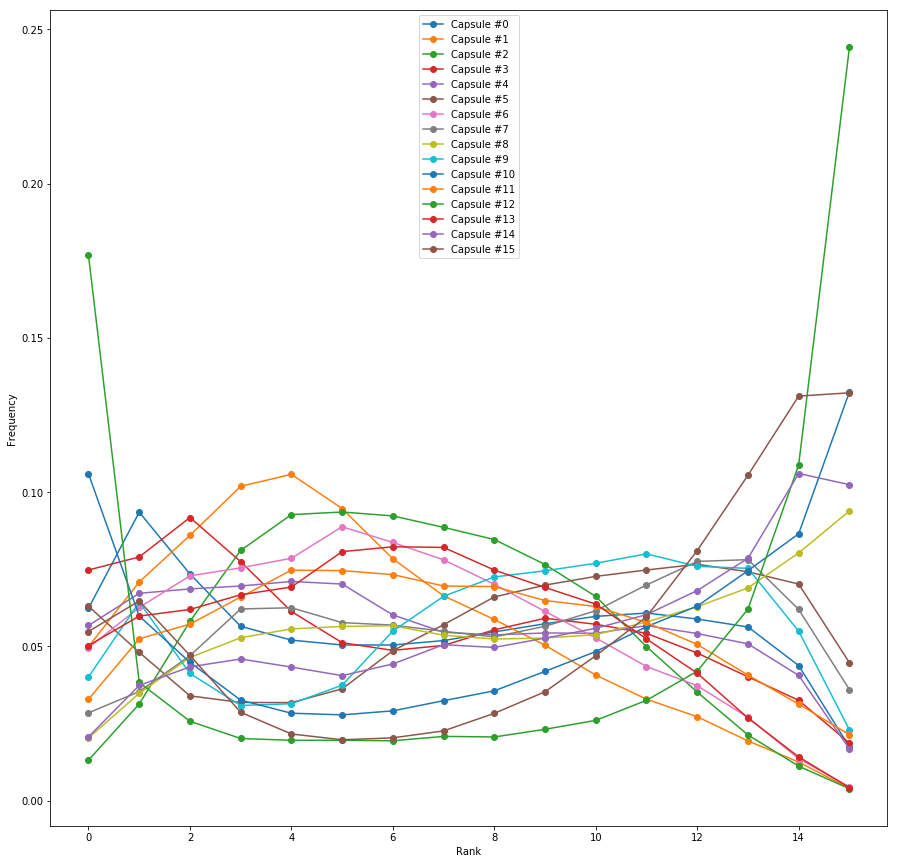}
  \caption{ Capsule mean rank-frequency over 10000 images. Each line-series shows the frequency with which a single capsule achieves each rank. Capsules are boosted to target a rank-0 frequency range of 0.04 to 0.1. Sparse training ensures that capsules specialize to each represent a range of related features found in the digit images.}\label{fig:5}
\end{figure}

\begin{figure}  
  \centering
    \includegraphics[width=0.45\textwidth]{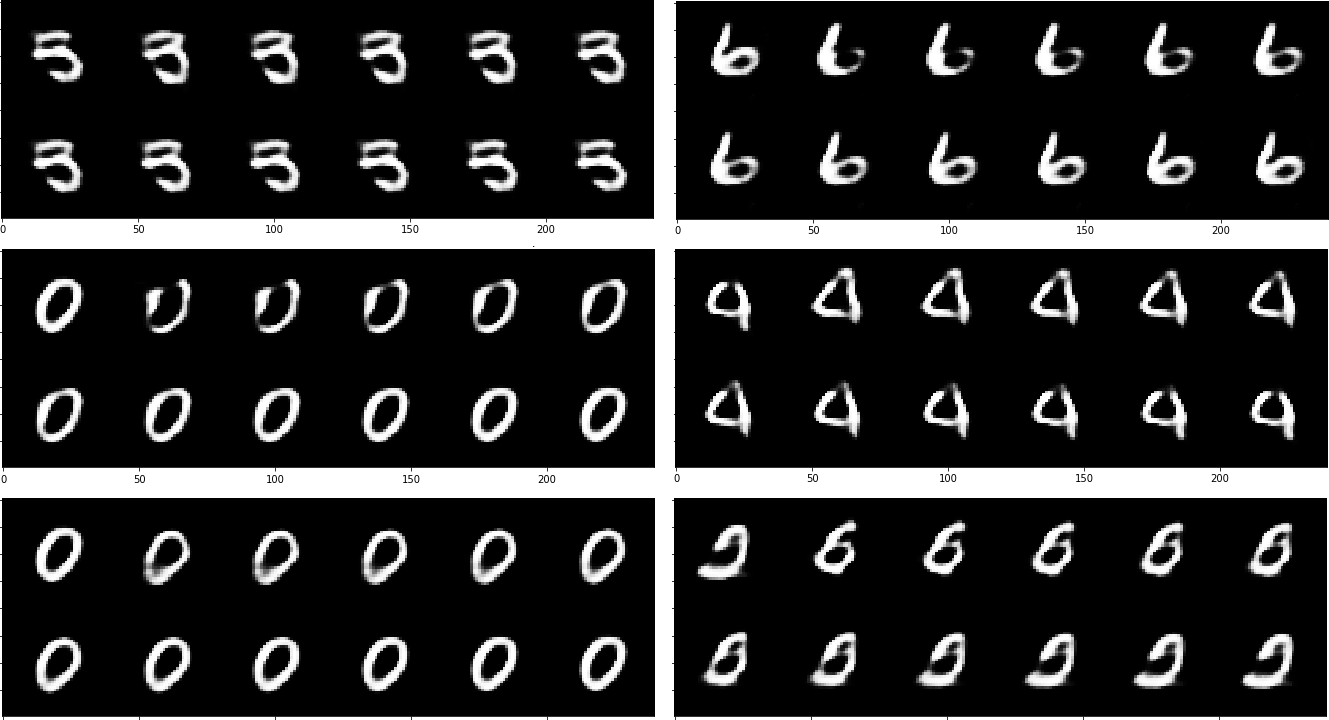}
  \caption{Example equivariance observed with sparse unsupervised training. Each panel is produced by varying the output of a single capsule parameter at intervals of $0.2$ between $-1$ and $1$. Digit form variations include pinching, skewing, morphing between digits (e.g. 3-5 top-left, and 6-2 lower-right), loop closure or completion, and stroke thickness or intensity.}\label{fig:6}
\end{figure}

The frequency of each latent capsule $j$ achieving rank 0 is computed exactly within each batch and measured as an exponentially weighted moving average over many batches with decay $\alpha=0.99$. This is heavily influenced by recent samples but was sufficiently accurate for our experiments. In our model, no hyperparameters vary over time but capsule statistics are updated every $N=50$ batches. Figure \ref{fig:5} shows the frequency with which each latent capsule achieves each rank over 10,000 affNIST test images, given sparse unsupervised training. As intended, we observe that latent capsules are active with a range of frequencies around the target range ($\mu_{min}=0.04$,  $\mu_{max}=0.1$) with a boost update $d=0.1$. Initial boost value for each latent capsule $\mathbf{g}_j=1$. 

This demonstrates that the boosting algorithm is working effectively. Some digit forms are observed more frequently than others, so a range of frequencies is expected. Intuitively, we wish to allow the data to determine latent capsule frequency as long as all available capsules are useful on occasion. 

Note that in SPARSECAPS latent capsule boosting and masking is only applied to reconstruction output and only affects capsule training. It is possible that boosting capsule coefficients during the routing iterations would be beneficial.

\subsubsection{Per-Image}
The $rank$ function (equation \ref{eq:2}) returns the ascending rank of the latent capsule $j$ in batch samples $k$. Parameter $\gamma$ determines the sparsity of the output. $v'$ (equation \ref{eq:5}) replaces $v$ as the output of the latent capsules layer used for classification and reconstruction.

\begin{equation}
\label{eq:1}
z_{jk} = \psi_{jk} \cdot \mathbf{g}_j
\end{equation}

\begin{equation}
\label{eq:2}
r_{jk} = rank( z_{k}, j )
\end{equation}


\begin{equation}
\label{eq:4}
m_{jk} = e^{-\gamma \frac{r_{jk}}{L-1}}
\end{equation}

\begin{equation}
\label{eq:5}
v' = v \circ m
\end{equation}

\subsubsection{Per-Batch}
After each batch, we update the online statistics about latent capsule utilization (equations \ref{eq:6} and \ref{eq:7}). Every $N$ batches we adjust capsule boost values $\mathbf{g}$ (see procedure \ref{alg:1}).

\begin{equation}
\label{eq:6}
\mathbf{\epsilon}_{j} = \frac{ 1}{K \cdot J} \cdot \sum_k{ 
\left\{\!\begin{aligned}
1,& \text{ if } r_{jk} = 0\\ 
0,& \text{ otherwise}
\end{aligned}\right\} }
\end{equation}

\begin{equation}
\label{eq:7}
\alpha = 0.99
\end{equation}

\begin{equation}
\label{eq:8}
\mathbf{\mu}_{j}' = \alpha \cdot \mathbf{\mu}_j + (1 - \alpha) \cdot \mathbf{\epsilon}_{j}
\end{equation}


\begin{algorithm}
\caption{Post-batch capsule boost update. $n$ is the current batch count.} \label{alg:1}
\begin{algorithmic}
\STATE $d = 0.1$
\STATE $N = 50$
\STATE $n \gets n +1$
\IF {($n \bmod N) <> 0 $} \RETURN
\ENDIF
\FOR{\texttt{each latent capsule $j$}}
\IF {$\mathbf{\mu}_{j} < \mu_{min}$} 
\STATE {$\mathbf{g}_{j} = \mathbf{g}_{j} + d$}
\ENDIF
\IF {$\mathbf{\mu}_{j} > \mu_{max}$} 
\STATE {$\mathbf{g}_{j} = \max( 1, \mathbf{g}_j - d )$}
\ENDIF
\ENDFOR
\end{algorithmic}
\end{algorithm}

\section{Results \& Discussion}
It is desirable to understand the utility of the unsupervised SPARSECAPS representation in discriminatory tasks such as image classification, and to compare the qualities of SPARSECAPS and SUPCAPS models. For this purpose we added a supervised SVM layer with a Radial Basis Function kernel to the output of the pre-trained SPARSECAPS. We will refer to the SVM as `the classification layer'.

In all results, SPARSECAPS was trained for 30,000 steps (each step is a mini-batch of 128 images). 

\subsection{Generalization to affNIST}
An important question is whether the learnt unsupervised representation can generalize to the affNIST dataset, in which affine transformations have been applied to MNIST digits. SUPCAPS achieved 79\% on this task. SPARSECAPS achieved 90\% (but see notes below that qualify this result). 

SUPCAPS was trained only on the MNIST dataset. In all experiments, the SPARSECAPS network was also trained only on the MNIST dataset similar to \cite{Sabour2017}.

We report 2 sets of supervised classification results: (a) the classification layer was trained on SPARSECAPS’ affNIST output, and tested on SPARSECAPS output on a disjoint affNIST set. (b) the classification layer was trained on SPARSECAPS’ MNIST output, and tested on an affNIST set. The two results answer different questions. Results (a) show whether the unsupervised representation trained on MNIST was able to generalize to represent the transformed digits of affNIST. The classifier layer is not expected to be able to generalize and therefore needs to observe the range of outputs from SPARSECAPS when exposed to affNIST data. If SPARSECAPS successfully learns an equivariant representation, we expect its output to vary when digits are viewed from different perspectives (i.e. after affNIST transformation). Conversely, an invariant representation would not vary in this way. Successful generalization from MNIST training images to affNIST test images suggests that SPARSECAPS has learned a representation that can represent affine transformations.

\begin{table}[h]
\begin{center}
\begin{tabular}{ | p{4.6cm}| p{1.2cm}  | p{1.2cm} | }
\hline
Algorithm & MNIST & affNIST \\ \hline
Conventional ConvNet \cite{Sabour2017} & 99.22\% & 66\% \\ \hline
SUPCAPS \cite{Sabour2017} & 99.23\% & 79\% \\ \hline
SPARSECAPS + SVM (a) & 99\% & \textbf{90.12\%} \\ \hline
SPARSECAPS + SVM (b) & 99\% & 16.99\% \\ \hline
Unmasked SPARSECAPS + SVM (a) & 97.37\% & 60.63\% \\ \hline
Unmasked SPARSECAPS + SVM (b) & 97.37\% & 15.70\% \\ \hline
\end{tabular}
\caption[Table caption text]{MNIST and affNIST classification accuracy for algorithms trained only on MNIST. To achieve a high affNIST score, algorithms must generalize from MNIST to affNIST. Test conditions (a) SVM trained on SPARSECAPS' affNIST output (b) SVM trained on SPARSECAPS' MNIST output.}
\label{table:generalize}
\end{center}
\end{table}

For comparison, results (b) show performance when the classifier layer is only allowed to observe SPARSECAPS output in response to MNIST images. Unsurprisingly, classification accuracy is poor. Good MNIST accuracy does not generalize to good affNIST accuracy. Neither result is exactly equivalent to the supervised training and testing of SUPCAPS. 

To demonstrate that sparse training of capsules is necessary, Table \ref{table:generalize} includes the result for our initial naive implementation of unsupervised capsules without sparse masking, trained on MNIST. In addition to the observed loss of equivariances, the unsupervised representation learnt does not generalize from MNIST to affNIST even when the classifier layer is trained on affNIST output. This suggests that capsular qualities are necessary for generalization from MNIST to affNIST.

\subsection{affNIST Benchmark}
How does SPARSECAPS compare to conventional networks trained on affNIST? affNIST is not thoroughly studied and we could not find many results for comparison. Available results show that SPARSECAPS trained only on MNIST is similar or better than state-of-the-art for deep convolutional networks trained and tested on affNIST (Table \ref{table:benchmark}). The only result that is comparable to SPARSECAPS is \cite{Chang2017}, a complex custom CNN trained on MNIST data with scale and translation (in our work only translations are varied at training time).

\begin{table}[h]
\begin{center}
\begin{tabular}{ | p{2.4cm} | p{1.3cm} | p{1.5cm} | p{1.3cm} |}
\hline
Reference & MNIST & Trained on affNIST? & affNIST \\ \hline
\cite{Zhao2017} & 97.82\% & Yes & 86.79\%    \\ \hline
\cite{Chang2017} & 97.5\% & Mix* & \textbf{91.6\%}    \\ \hline
\cite{Shuhei2018} & 99.25\% & Yes & 87.55\%   \\ \hline
CNN in \cite{Sabour2017} & 99.22\% & No & 66\%   \\ \hline
SPARSECAPS & 99\% & No & \textbf{90.12\%}    \\ \hline
\end{tabular}
\caption[Table caption text]{MNIST and affNIST classification accuracy. SPARSECAPS shows that capsules networks performance is equal or better than most current convolutional networks, even when SPARSECAPS is trained on MNIST only and the convolutional networks are trained on affNIST. Note that preprocessing techniques and test conditions for MNIST may vary slightly between authors.\break*Chang et al. trained their affNIST network on an MNIST dataset with scale and translation variation but not full affine transformation.}
\label{table:benchmark}
\end{center}
\end{table}

\subsection{Equivariances in Sparse Latent Capsules}
We confirmed and visualized equivariances learned by SPARSECAPS by inspecting the reconstruction output while varying one parameter of the most active capsule after routing. As seen in \cite{Sabour2017}, reconstructed digit forms vary continuously in ways that plausibly represent handwriting variations. Since our latent-capsules do not have the same predefined meaning as SUPCAPS’ digit-caps, the equivariances discovered are not the same (see Figure \ref{fig:6}). Here, in addition to pinching and skewing of digits and stroke width variation, we also see closing and opening of the stroke-loops (e.g. digits zero and eight), mutations between similar digit-forms and other changes.

This is expected because unsupervised learning does not distinguish digits by class, only by appearance. In handwriting, each digit has at least one form, and typically 2 or 3 major forms in which it can appear. In some ambiguous cases, digit forms blur into one another. In the latter case, it is reasonable for a capsule to represent forms shared by 2 or more digits.

\subsection{Equivariances in Dense Latent Capsules}
Without sparsening, all parameters of each capsule make only minor edits to the reconstructed image (Figure \ref{fig:2}). The edits are scattered around the digit without any spatial correlation. This illustrates the degenerate case where a capsules network is trained in unsupervised fashion without sparsening. We conclude that equivariances are have not been learned.

\subsection{Other Metrics}
MNIST classification accuracy at 99\% was not significantly affected by the change from supervised to sparse unsupervised capsules with an SVM classifier. We speculate that the number of parameters in the network is easily sufficient to classify that dataset and the bottleneck is not too tight. Reconstruction loss is good; batch mean square error is 7.75 and 18.49 on MNIST and affNIST datasets respectively. 

Figure \ref{fig:4} compares routing coefficients between SUPCAPS, dense unsupervised capsules and SPARSECAPS, illustrating the effect of capsule specialization on dynamic routing behaviour. It confirms that with sparse training, dynamic routing is again able to selectively activate relevant capsules.

\section{Conclusions}
We discovered that without supervised masking of latent capsule layers to impose a specialized identity on each latent capsule, the desirable characteristics of capsules networks are lost. This suggests that deeper capsules networks, lacking such masking in hidden layers, will not work as intended. The failure mode for these networks is to behave as a convolutional autoencoder, so this is not immediately obvious. Although we demonstrated these problems while using SUPCAPS’ dynamic routing, it is likely the same issues occur when using capsules with EM-routing \cite{Hinton2018}, although perhaps to a lesser degree due to the cluster-finding behaviour of EM. Backpropagation of the supervised digit-class loss acts to sparsen the final capsules layer; without it, we expect that capsule activation will become more uniform with increasing depth, even with EM-routing.

We found that sparse unsupervised training of latent capsule layers both restored capsule qualities and in fact produced (arguably) superior generalization performance on the affNIST dataset. This suggests that sparse training could enable deeper capsule networks.

In future work we hope to integrate SPARSECAPS with EM-routing and test on the SmallNORB dataset \cite{LeCun2004}. This dataset tests 3D object recognition given pose and illumination changes. 

Although not tested, capsules with EM-routing appear to be relatively robust to adversarial examples. We expect that the same is true of SPARSECAPS due to the unsupervised training regime. This needs to be empirically confirmed.

The lifetime capsule frequency hyperparameters were derived by calculating the uniform frequency as a midpoint and then empirically optimizing the bounds. (In fact these were the only parameters empirically optimized for this paper). In future, it should be possible to make the frequency range a function of the number of capsules, or perhaps with online adaptation to ensure good resource utilization.

\section*{Acknowledgments}
We would like to thank \cite{Sabour2017} and other researchers who contribute to open research by  publicly releasing resources such as source code.

\subsection*{Author Contributions}
Abdelrahman Ahmed wrote the code, planned and executed the experiments, analyzed the results, and helped write the paper. David Rawlinson devised the sparse unsupervised capsules algorithm, supervised the experiments, and drafted the paper. Gideon Kowadlo conceived the sparse unsupervised capsules research, supervised the experiments and helped write the paper. 

\bibliography{sparsecaps.bib}
\bibliographystyle{icml2018}

\clearpage

\appendix
\section{Appendix}
\setcounter{figure}{0}
\renewcommand{\thefigure}{S\arabic{figure}}

\subsection{Source Code}
Full source code to reproduce these results can be found in our GitHub repository at:

\url{https://github.com/ProjectAGI/sparse-unsupervised-capsules}

Our repository is a fork of the SUPCAPS code from \cite{Sabour2017}. The original code can be found at: 

\url{https://github.com/Sarasra/models/tree/master/research/capsules}

\end{document}